\documentclass[runningheads]{template/llncs}

\usepackage{graphicx}
\usepackage{multirow}
\usepackage{adjustbox}
\usepackage[acronym]{glossaries}
\usepackage{wrapfig}
\usepackage{caption}
\usepackage[format=hang, justification=RaggedRight]{subcaption}
\usepackage{amssymb}

\let\subparagraph\paragraph
\usepackage[pagestyles]{titlesec}
\usepackage{layouts}
\usepackage[noend, linesnumbered, boxed]{algorithm2e}
\usepackage{floatrow}
\usepackage[skins]{tcolorbox}
\usepackage{mathtools}
\usepackage{hyperref}

\newacronym{DML}{DML}{\textit{deep metric learning}}
\newacronym{LEV}{LEV}{\textit{linguistic embedding vector}}
\newacronym{BFS}{BFS}{\textit{Bayes factor scoring}}
\newacronym{UAL}{UAL}{\textit{uncertainty adaptation layer}}
\newacronym{AV}{AV}{authorship verification}
\newacronym{LEVs}{LEVs}{\textit{linguistic embedding vectors}}
\newacronym{O2D2}{O2D2}{\textit{out-of-domain detection}}
\newacronym{ECE}{\texttt{ECE}}{\textit{expected calibration error}}
\newacronym{MCE}{\texttt{MCE}}{\textit{maximum calibration error}}

\newcommand{\loss}{\ensuremath{\mathcal{L}}}
\DeclarePairedDelimiter{\ceil}{\lceil}{\rceil}

\titlespacing*{\section}{0pt}{1.8ex}{.8ex}
\titlespacing*{\subsection}{0pt}{1.6ex}{.5ex}
\titlespacing*{\subsubsection}{0pt}{1.6ex}{.5ex}

\begin{document}

\setlength{\textfloatsep}{8pt}
\setlength{\abovedisplayskip}{5.pt}
\setlength{\belowdisplayskip}{6.pt}
\setlength{\abovedisplayshortskip}{5.pt}
\setlength{\belowdisplayshortskip}{6.pt}

\title{Self-Calibrating Neural-Probabilistic Model for Authorship Verification Under Covariate Shift}
\titlerunning{ }

%
\author{Benedikt Boenninghoff\inst{1} \and
Dorothea Kolossa\inst{1} \and
Robert M. Nickel\inst{2} }
\authorrunning{ }

%
\institute{Ruhr University Bochum, Germay \\
\email{\{benedikt.boenninghoff, dorothea.kolossa\}@rub.de} 
\and
Bucknell University, USA \\
\email{rmn009@bucknell.edu}
}
\maketitle              
\begin{abstract}
We are addressing two fundamental problems in authorship verification (AV): Topic variability and miscalibration. Variations in the topic of two disputed texts are a major cause of error for most AV systems.
In addition, it is observed that the underlying probability estimates produced by deep learning AV mechanisms oftentimes do not match the actual case counts in the respective training data. As such, probability estimates are poorly calibrated. We are expanding our framework from PAN 2020 to include Bayes factor scoring (BFS) and an uncertainty adaptation layer (UAL) to address both problems. Experiments with the 2020/21 PAN AV shared task data show that the proposed method significantly reduces sensitivities to topical variations and significantly improves the system's calibration.


\end{abstract}
    \section{Introduction}
\begin{figure}[t]
\centering
\includegraphics[width=0.9\textwidth]{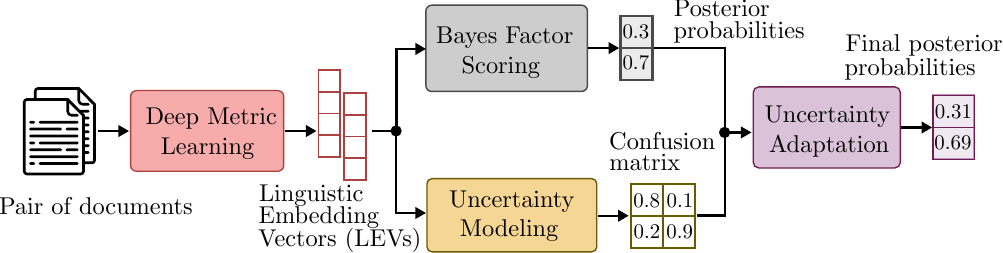}
\vspace{-0.2cm}
\caption{Our proposed end-to-end neural-probabilistic model.} 
\label{fig:model}
\end{figure}

 Computational \gls{AV} is often described as the task to automatically accept or reject the identity claim of an {\em unknown\/} author by comparing a disputed document with a reference document written by
 a {\em known\/} author.
 \gls{AV} can be described mathematically as follows.
 Suppose we have a pair of documents $\mathcal{D}_1$ and $\mathcal{D}_2$ with an associated ground-truth hypothesis $\mathcal{H}_a$ for $a\in\{0,1\}$.
 The value of $a$ indicates if the two documents were written by the same author ($a=1$) or by different authors ($a=0$).
Automated systems usually calculate scores or likelihood ratios to distinguish between the same-author and the different-authors cases. The score-based task can formally be expressed as a mapping
$f\!\!:\!\!\{\mathcal{D}_1, \mathcal{D}_2\} \longrightarrow
s \in [0,1]$. 
Usually, the estimated label $\widehat{a}$ is obtained from a threshold test applied to the score/prediction value $s$. For instance, we may choose $\widehat{a}=1$ if $s>0.5$ and $\widehat{a}=0$ if $s<0.5$. The PAN 2020/21 shared tasks also permit the return of a \textit{non-response} (in addition to $\widehat{a}=1$ and $\widehat{a}=0$)
in cases of high uncertainty~\cite{kestemont:2020}, e.g.~when $s$ is close to 0.5.

The current PAN AV challenge moved from a closed-set task in the previous year to an open-set task in 2021, i.e.~a scenario in which the testing data contains \textit{only} authors and topics that were \textit{not} included in the training data. We thus expect a \textit{covariate shift} between training and testing data, i.e.~the distribution of the features extracted from the training data is expected to be different from the distribution of the testing data features.
It was implicitly shown in~\cite{kestemont:2020} that such a covariate shift, due to topic variability, is a major cause of errors in authorship analysis applications.

Our proposed framework\footnote{The source code is accessible online: \url{https://github.com/boenninghoff/pan\_2020\_2021\_authorship\_verification}} is presented in~Fig.~\ref{fig:model}.
In~\cite{boenninghoff:2020a}, we introduced the concept of \gls{LEVs}, where we perform \gls{DML} to encode the stylistic characteristics of a pair of documents into a pair of fixed-length representations, $\boldsymbol{y}_i$ with $i\in\{1, 2\}$.
Given the \gls{LEVs}, a \gls{BFS} layer computes the posterior probability for a trial. Finally, we propose an \gls{UAL} including uncertainty modeling and adaptation to correct possible misclassifications and to return corrected and calibrated posteriors, $p(\mathcal{H}_{\widehat{a}} | \boldsymbol{y}_1,  \boldsymbol{y}_2)$ with $\widehat{a}\in \{0, 1\}$.

For the decision, whether to accept $\mathcal{H}_0$/$\mathcal{H}_1$ or to return a non-response, it is desirable that the concrete value or outcome of the posterior $p(\mathcal{H}_{\widehat{a}} | \boldsymbol{y}_1,  \boldsymbol{y}_2) = s$ has a reliable \textit{confidence} score. Ideally, this confidence score should match the true probability of a correct outcome. 
Following~\cite{pampari2020unsupervised}, our neural-probabilistic model is said to be well-calibrated if its  posterior probabilities  match  the  corresponding  empirical  frequencies.
Inspired by~\cite{pmlr-v70-guo17a}, we take up the topic of calibration of confidence scores in the field of deep learning to the case of binary \gls{AV}.
A perfectly calibrated authorship verification system can be defined as
\begin{align}
\label{eqn:calibration}
\mathbb{P}\bigg(\mathcal{H}_{\widehat{a}} = \mathcal{H}_a \bigg| p(  \mathcal{H}_{\widehat{a}} | \boldsymbol{y}_1,  \boldsymbol{y}_2) = s \bigg) = s 
\quad \forall s  \in [0,1], ~~a\in\{0, 1\},~~\widehat{a} \in\{0, 1\}.
\end{align}
\indent As mentioned in~\cite{pmlr-v70-guo17a} we are not able to directly measure the probability in Eq.~\eqref{eqn:calibration} and the authors proposed two empirical approximations, i.e. 
the~\gls{ECE} and the \gls{MCE} to capture the miscalibration of neural networks.

\begin{figure}[t]
\vspace*{-0.4cm}
\centering
\begin{subfigure}[t]{0.42\textwidth}
\centering
\includegraphics[width=0.9\textwidth]{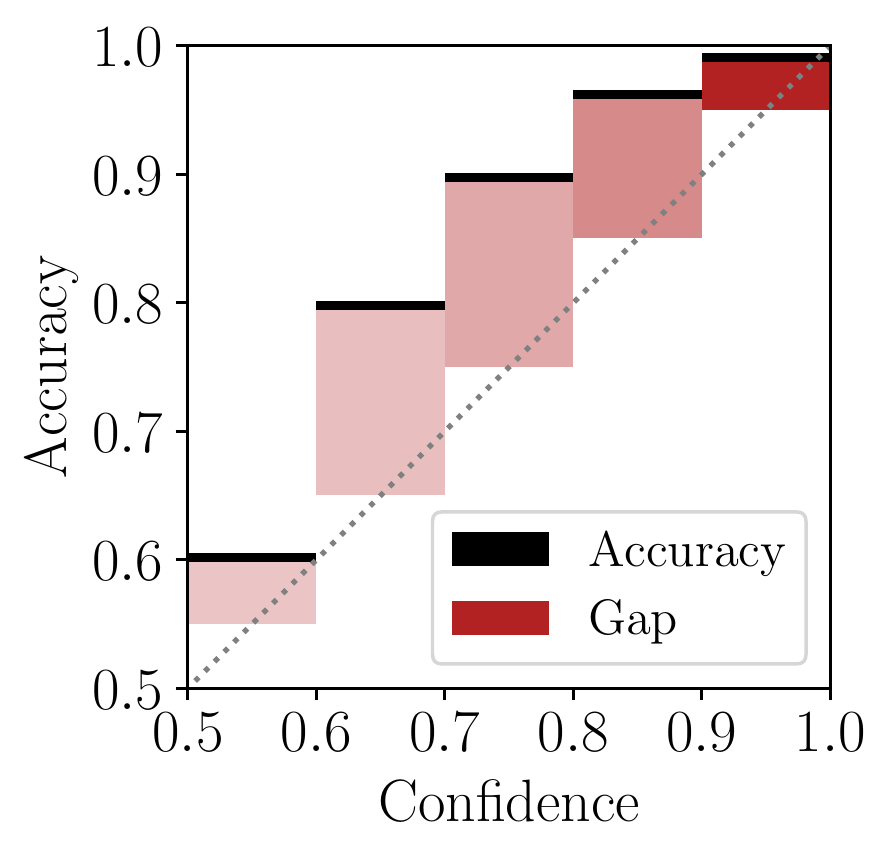}
\vspace*{-0.2cm}
\caption{Uncalibrated confidence values of our PAN 2020 submission.}
\end{subfigure}
\quad\quad
\begin{subfigure}[t]{0.42\textwidth}
\centering
\includegraphics[width=0.9\textwidth]{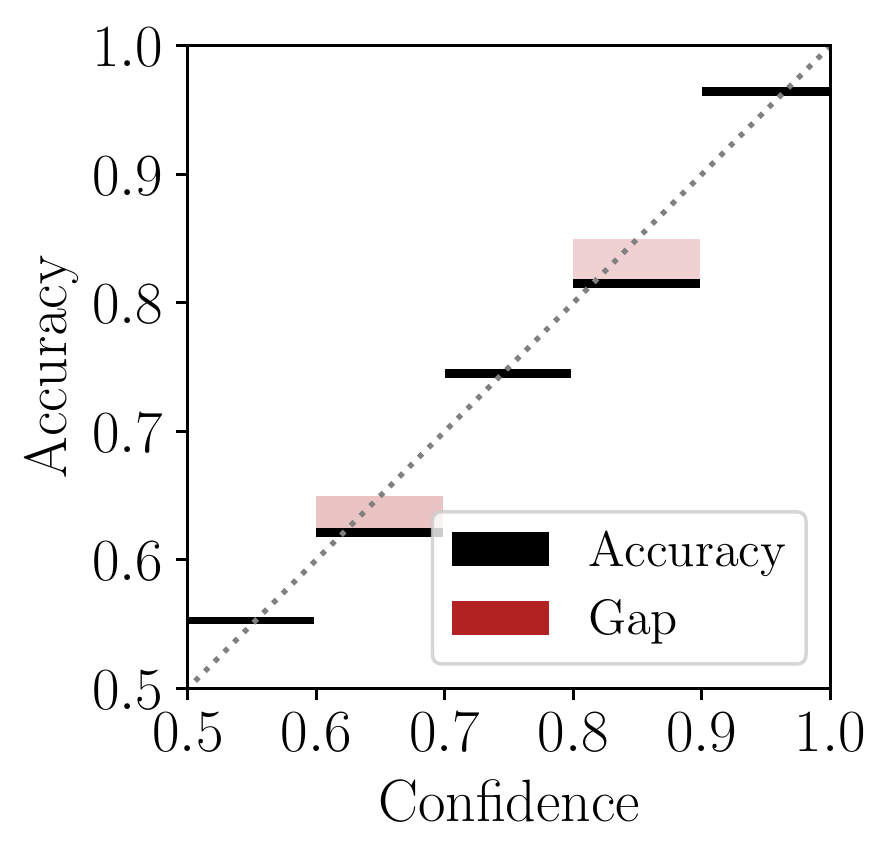}
\vspace*{-0.2cm}
\caption{Calibrated confidence values of our PAN 2021 submission.}
\end{subfigure}
\vspace*{-0.2cm}
\caption{Reliability diagrams for our PAN 2020 submission and the proposed 2021 submission. The red bars are darker for bins with a higher number of trials.} 
\label{fig:rel2021}
\end{figure}

Another way to visualize how well our model is calibrated, is to draw the \text{reliability  diagram}.
The confidence interval is discretized into a fixed number of bins. Afterwards, we compute the average confidence and the corresponding accuracy for each bin.
Fig.~\ref{fig:rel2021} illustrates the differences between our PAN 2020 submission and our extended version for PAN 2021.
The red gaps between accuracy and confidence in Fig.~\ref{fig:rel2021}(a) indicate a miscalibration,
meaning that the system delivers \textit{under-confident} predictions since the accuracy is always larger than the confidence - resulting in a higher occurrence of false negatives. 
In contrast, the diagram in Fig.~\ref{fig:rel2021}(b) shows that, for our proposed new method, accuracy and confidence are closer, if not equal.

In this work, we expand our method from PAN 2020 by adding new system components, evaluate performance w.r.t.~\textit{authorship} and \textit{topical label}, and illustrate the effect of the fine-tuning of some core hyper-parameters.
Our experiments show that, even though we are not able to fully suppress the misleading influence of topical variations, we are at least able to reduce their biasing effect.
 
\section{Text Preprocessing Strategies}
\label{sec:TPPS}

\subsection{PAN2020 Dataset Split}
The \textit{fanfictional} dataset for the PAN 2020/21 \gls{AV} tasks is described in~\cite{kestemont:2020} and contains 494,227 unique documents written by 278,162 unique authors, grouped into 1,600 unique fandoms.
We split the dataset into two disjoint (w.r.t.~authorship and fandom) datasets and removed all overlapping documents. The training set contains 303,142 documents of 1,200 fandoms written by 200,732 authors.
The test set has 96,027 documents of 400 fandoms written by 77,430 authors.

\subsection{Re-sampling Document Pairs}

\begin{figure}[t]
\vspace*{-0.4cm}
\begin{floatrow}
\hspace*{-.45cm}
\ffigbox[1.45\linewidth]{
\begin{tcolorbox}[blanker,width=(\linewidth-.8cm)]
\scalebox{0.8}{
\begin{algorithm}[H]
	\DontPrintSemicolon
	\small
	\textbf{Input:} Sorted documents w.r.t authorship and fandom\;
	\textbf{Output:} Pairs of documents\;
	\BlankLine
    \While{author with a document is available}{
	    \For{all authors}{	
		\uIf{$r\sim \mathcal{U}[0, 1] < \delta_1$}{
		    \uIf{$r\sim \mathcal{U}[0, 1] < \delta_2$}{
    			Try to sample a \texttt{SA\_SF} pair\;
    		}
    		\uElse{
               Try to sample a \texttt{SA\_DF} pair\;
    		}
    	}\uElse{
               Try to sample a \texttt{DA} candidate
    		}
	}
	Delete author if all documents are sampled\;
	}
    \While{two documents are available}{
	   		    \uIf{$r\sim \mathcal{U}[0, 1] < \delta_3$}{
    			Try to sample a \texttt{DA\_SF} pair\;
    		}
    		\uElse{
               Try to sample a \texttt{DA\_DF} pair\;
    		}
	}
\end{algorithm}
}
\end{tcolorbox}
}
{\caption{Pair re-sampling procedure.~~~~~~~~~~~~~~~~~~}
\label{fig:resample}
}
\hspace*{-2.4cm}
\ffigbox[\linewidth]{
\includegraphics[width=0.45\textwidth]{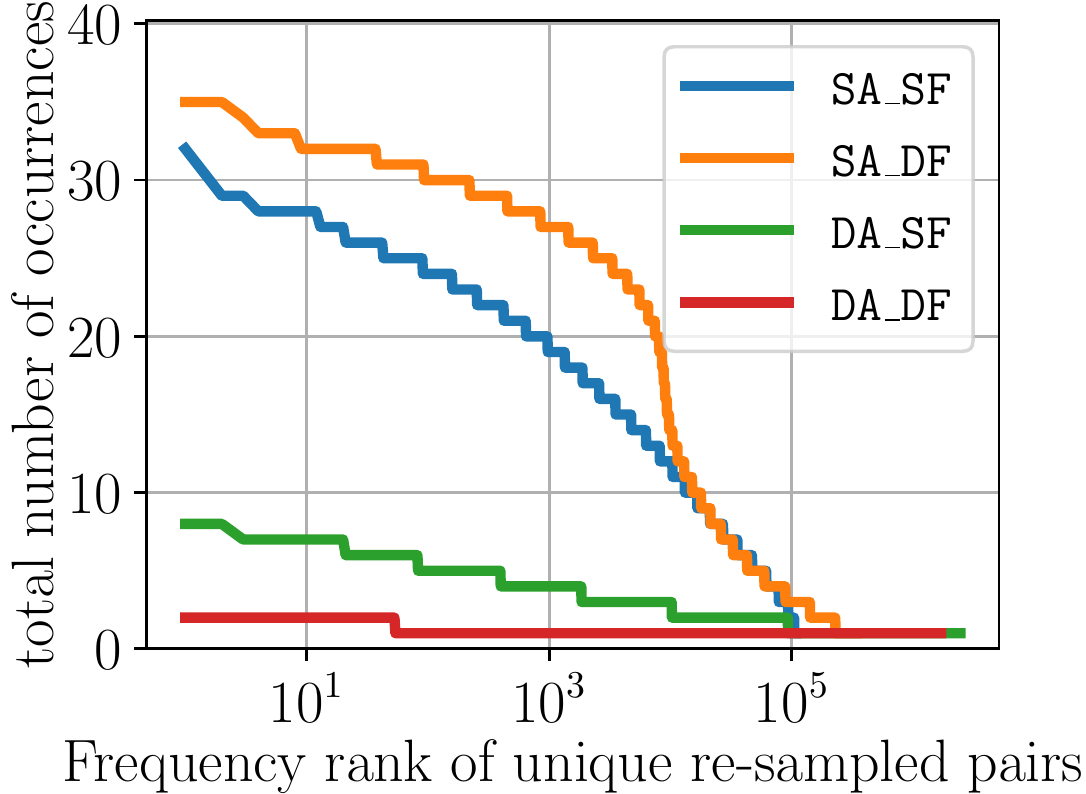}
}
{\caption{~Zipf plot of the pair counts.}
\label{fig:zipf}}
\end{floatrow}
\end{figure}

The size of the training set can be augmented by re-sampling new document pairs in each epoch, as illustrated in the pseudo-algorithm in Fig.~\ref{fig:resample}. Each document pair is characterized by a tuple $(a, f)$, where $a\in \{0, 1\}$ denotes the authorship label and $f\in \{0, 1\}$ describes the equivalent for the fandom. Each document pair is assigned to one of the subsets\footnote{\texttt{SA}=same author, \texttt{DA}=different authors,
\texttt{SF}=same fandom, \texttt{DF}=different fandoms}
\texttt{SA\_SF}, \texttt{SA\_DF}, \texttt{DA\_SF},  and \texttt{DA\_DF}
in correspondence with its label tuple $(a,f)$.

The algorithm in Fig.~\ref{fig:resample} follows three constraints: Firstly, all documents contribute equally to the neural network training in each epoch.
Secondly, repetitively re-sampling of the same document pairs should be reduced. Thirdly, each document should appear in equal numbers in all subsets. 
Our re-sampling strategy roughly consists of two while loops, where $\mathcal{U}[0,1]$ represents a uniform sampler over the half-open interval $[0,1)$. In the first while loop (lines 3-12) we iterate over all authors until all documents have been sampled either to one of the same-author sets (\texttt{SA\_SF} and \texttt{SA\_DF}) or have been chosen to be a \texttt{DA} candidate. In the second while loop (lines 13-17), we take all collected \texttt{DA} candidates to sample \texttt{DA\_SF} and \texttt{DA\_DF} pairs. The parameters $\delta_1$, $\delta_2$, and $\delta_3$ control the distributions of the subsets. 
We chose $\delta_1=0.7$, $\delta_2=0.6$ and $\delta_3=0.6$.
As a result, the epoch-wise training sets are not balanced, rather, we obtain approximately $70$\% different-authors pairs and $30$\% same-author pairs. 

Fig.~\ref{fig:zipf} shows a Zipf plot of the pair counts. It can be seen that there is still a high repetition regarding the same-author pairs since each author generally contributes only with a small number of documents. 

During the evaluation stage, the verification task is performed on the test set only. 
The pairs of the test set are sampled once and then kept fixed.
In Section~\ref{sec:EXP}, we briefly report on the system performance for all subsets and then proceed with the analysis of a more challenging case: We removed all \texttt{SA\_SF} and \texttt{DA\_DF} pairs. Finally, we have $5,216$ \texttt{SA\_DF} pairs and $7,041$ \texttt{DA\_SF} pairs, resulting in a nearly balanced dataset of $12,257$ test pairs.

\subsection{Sliding Windowing}

\begin{figure}[t]
\vspace*{-0.4cm}
 \centerline{\includegraphics[width=.95\textwidth]{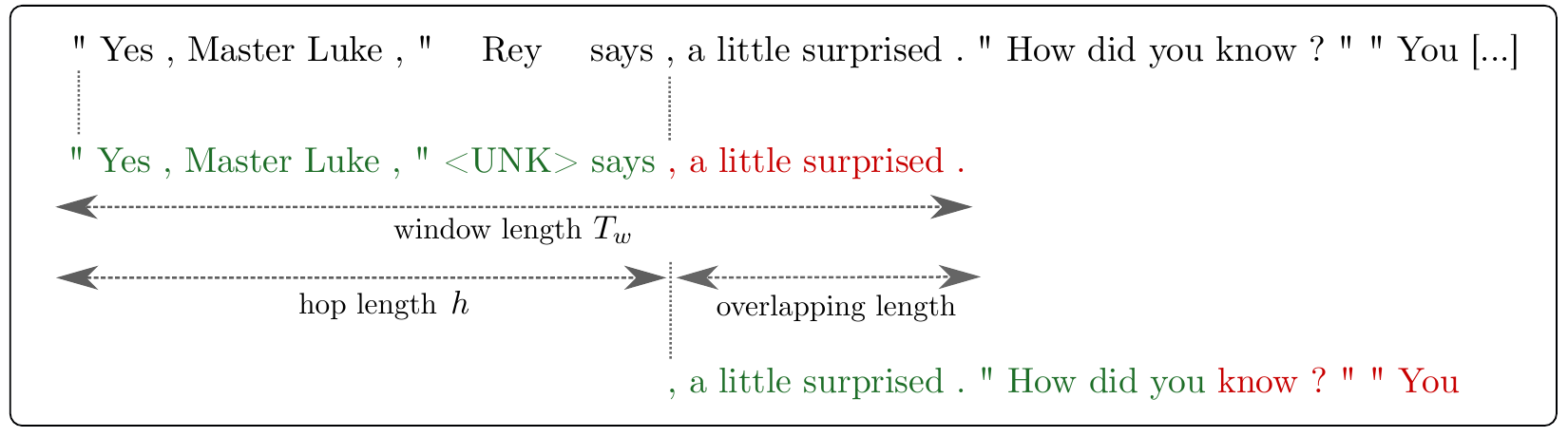}}
 \vspace*{-0.2cm}
 \caption{Example of our sliding window approach.} 
 \label{fig:sliwin}
 \end{figure}

As suggested in~\cite{boenninghoff:2020a}, we perform tokenization and generate \textit{sentence-like units} via a sliding window technique. An example is given in Fig.~\ref{fig:sliwin}. 
With the sliding window technique we obtain a compact representation for each document, where zero-padding tokens only need to be added to the last sentence. 
Given the total number of tokens per sentence $T_w$, the hop length $h$ and the total number of tokens $N$, the total number of sentence units per document is 
$T_s = \ceil[\big]{\frac{N - T_w + h}{h}}$.
We choose $T_w=30$ and $h=26$. The maximum number of sentence units per document is upper bounded by the GPU memory and set to $T_s=210$.

\subsection{Word Embeddings and Topic Masking}
\label{sec:we_andtm}
A disadvantage of the sliding window approach is that our sentence-like units differ from common sentence structures required by modern contextualized word embedding frameworks. Hence, we decided to represent a token by two different context-independent representations which are learned during training. Firstly, we initialize semantic word representations from the pretrained FastText model~\cite{bojanowski2016enriching}.
Secondly, we encode new word representations based on characters
~\cite{boenninghoff:2019b}.
We further reduce the vocabulary size for tokens and characters by mapping all rare token/character types to a special \textit{unknown} (\texttt{<UNK>}) token which can be interpreted as a topic masking strategy~\cite{stamatatos-2017-authorship}. 
Finally we chose vocabulary sizes of $5,000$ tokens and $300$ characters. The embedding dimensions are given by $D_w=300$ for words and $D_c=10$ for characters.

\section{Neural-Probabilistic Model}
\label{sec:NPM}


\subsection{Neural Feature Extraction and Deep Metric Learning}
\label{seq:neural_dml}
Neural feature extraction and the deep metric learning are realized in the form of a \textit{Siamese} network, mapping both documents into neural features through exactly the same function.

\subsubsection{Neural Feature Extraction: }
After text preprocessing, a single document consists of a list of $T_s$ ordered sentences. Each sentence consists of an ordered list of $T_w$ tokens. Again, each token consists of an ordered list of $T_c$ characters.
As mentioned in Section~\ref{sec:we_andtm}, we implemented a characters-to-word encoding layer to obtain word representations. The dimension is set to $D_r=30$. 
The system passes a fusion of token and character embeddings into a two-tiered bidirectional LSTM network with attentions, 
\begin{align}
    \label{eq:docemb}
    \boldsymbol{x}_i = \text{NeuralFeatureExtraction}_{\boldsymbol{\theta}}\big(
        \boldsymbol{E}_i^w, 
        \boldsymbol{E}_i^c
    \big),
\end{align}
where $\boldsymbol{\theta}$ contains all trainable parameters, 
$\boldsymbol{E}^w_i \in \mathbb{R}^{T_s \times T_w \times D_w}$ represents word embeddings and $\boldsymbol{E}^c_i \in \mathbb{R}^{T_s \times T_w \times T_c  \times D_c}$ represents character embeddings.
A comprehensive description can be found in~\cite{boenninghoff:2019b}.

\subsubsection{Deep Metric Learning: }
We feed the document embeddings $\boldsymbol{x}_i$ in Eq.~\eqref{eq:docemb} into a metric learning layer,
\begin{align}
\boldsymbol{y}_i = \tanh\big(\boldsymbol{W}^{\textsc{DML}}  \boldsymbol{x}_i +  \boldsymbol{b}^{\textsc{DML}}\big),
\end{align}
which yields the two \gls{LEVs} $\boldsymbol{y}_1$ and $\boldsymbol{y}_2$ via the trainable parameters $\boldsymbol{\psi}=\{\boldsymbol{W}^{\textsc{DML}},$ $\boldsymbol{b}^{\textsc{DML}}\}$. 
We then compute the Euclidean distance between both LEVs,
\begin{align}
    \label{eq:ED}
    d(\boldsymbol{y}_1,\boldsymbol{y}_2) = \left\lVert \boldsymbol{y}_1 - \boldsymbol{y}_2 \right\rVert_2^2.
\end{align}

\subsubsection{Probabilistic contrastive loss: }
\begin{figure}[t]
\vspace*{-0.4cm}
\begin{floatrow}
\ffigbox[6.3cm]{
\includegraphics[width=0.52\textwidth]{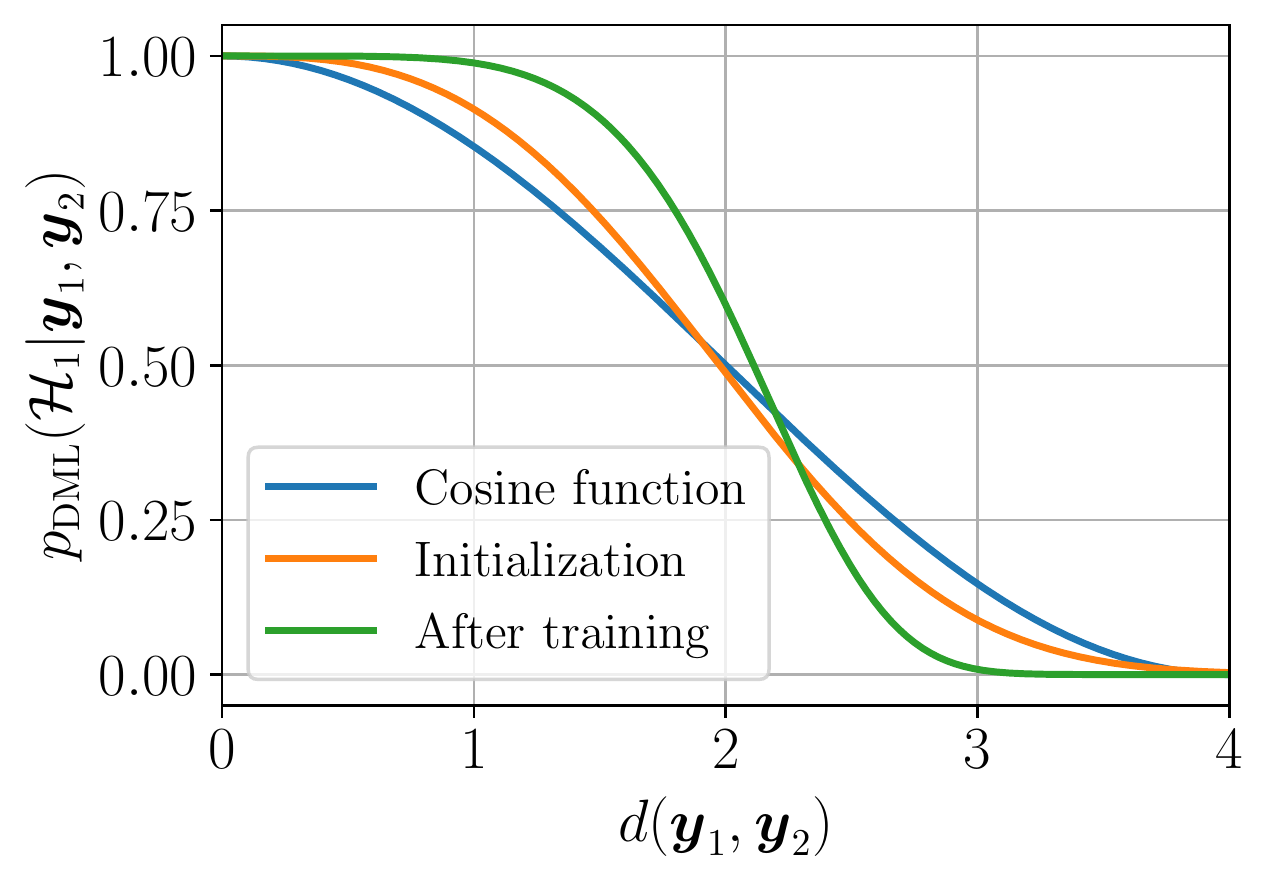}
}
{\caption{Learned mapping function of the probabilistic contrastive loss.}
\label{fig:mapping}
}
\hspace{.1cm}
\ffigbox[5.1cm]{
\includegraphics[width=0.4\textwidth]{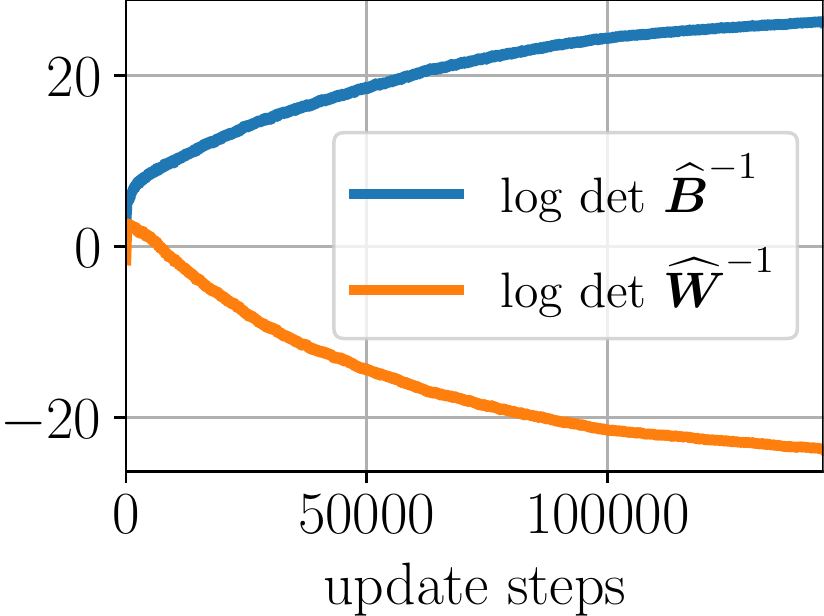}
}
{\caption{Entropy curves during training.}
\label{fig:loss}
}
\end{floatrow}
\end{figure}

In~\cite{boenninghoff:2019b}, we chose the modified contrastive loss,
\begin{align}
  \label{eq:lossdml1}
\loss^{\text{DML}}_{\boldsymbol{\theta}, \boldsymbol{\psi}} &= a \cdot \max \left\{ d(\boldsymbol{y}_1,\boldsymbol{y}_2) - \tau_s, 0 \right\}^2
 +       (1-a) \cdot \max \left\{ \tau_d - d(\boldsymbol{y}_1,\boldsymbol{y}_2) , 0\right\}^2,
\end{align}
with  $\tau_s=1$ and $\tau_d=3$. 
With the contrastive loss all distances between same-author pairs are forced to stay below $\tau_s$ and conversely, distances between different-authors pairs are forced to remain above $\tau_d$.
A drawback of this contrastive loss is that its output cannot be interpreted as a probability. We therefore introduce a new \textit{probabilistic} version of the contrastive loss: Given the Euclidean distance of the \gls{LEVs} in Eq.~\eqref{eq:ED}, we apply a kernel function 
\begin{align}
\label{eq:DMLl}
p_{\text{DML}}(\mathcal{H}_1 |\boldsymbol{y}_1,\boldsymbol{y}_2) = \exp\big(- \gamma~ d(\boldsymbol{y}_1,\boldsymbol{y}_2)^{\alpha} \big),
\end{align}
where $\gamma$ and $\alpha$ can be seen as both, hyper-parameters or trainable variables. The new loss then represents a slightly modified version of Eq.~\eqref{eq:lossdml1},
\begin{align}
 \begin{split}
 \label{eq:lossdml2}
\loss^{\text{DML}}_{\boldsymbol{\theta}, \boldsymbol{\psi}} &= a \cdot \max \big\{ \tau_s- p_{\text{DML}}(\mathcal{H}_1 |\boldsymbol{y}_1,\boldsymbol{y}_2), 0 \big\}^2
 \\ & \quad +       (1-a) \cdot \max \left\{ p_{\text{DML}}(\mathcal{H}_1 |\boldsymbol{y}_1,\boldsymbol{y}_2) - \tau_d, 0\right\}^2,
\end{split}
\end{align}
where we set $\tau_s = 0.91$ and $\tau_d=0.09$.
Fig.~\ref{fig:mapping} illustrates the decision mapping of the new loss, transforming the distance scores into probabilities. The cosine similarity is a widely used similarity measure in \gls{AV}~\cite{bevendorff-etal-2019-bias}. Hence, we initialized $\alpha$ and $\gamma$ to approximate the cosine function in the interval $[0,4]$ (blue curve), which was the operating interval in~\cite{boenninghoff:2019b}. During training, we optimized $\alpha$ and $\gamma$, resulting in the green curve. We will discuss the effect in Section~\ref{sec:EXP}.

\subsection{Deep Bayes Factor Scoring}
The idea of pairwise  Bayes factor scoring was originally proposed in~\cite{6466371}. In~\cite{boenninghoff:2020a}, we adapted the idea to the context of \gls{AV}.
We assume that the \gls{LEVs} in Eq.~\eqref{eq:ED} stem from a Gaussian generative model that can be decomposed as $\boldsymbol{y} = \boldsymbol{s} + \boldsymbol{n}$,
where $\boldsymbol{n}$ characterizes a noise term, caused by e.g.~topical variations.
We assume that the writing characteristics of the author, measured in the observed LEV $\boldsymbol{y}$, lie in a latent stylistic variable $\boldsymbol{s}$.
The probability density functions for $\boldsymbol{s}$ and $\boldsymbol{n}$ are given by Gaussian distributions,
$ p(\boldsymbol{s}) = \mathcal{N}(\boldsymbol{s} | \boldsymbol{\mu}, \boldsymbol{B}^{-1})$
and
$p(\boldsymbol{n}) = \mathcal{N}(\boldsymbol{n} | \boldsymbol{0}, \boldsymbol{W}^{-1})$,
where $\boldsymbol{B}^{-1}$ defines the \textit{between-author} covariance matrix and $\boldsymbol{W}^{-1}$ denotes the \textit{within-author} covariance matrix.
We outlined in~\cite{boenninghoff:2020a} how to compute 
the likelihoods for both hypotheses.
The verification score for a trial is then given by the log-likelihood ratio: $ \text{score}(\boldsymbol{y}_1, \boldsymbol{y}_2) 
 = \log p(\boldsymbol{y}_1, \boldsymbol{y}_2 |\mathcal{H}_1) - \log p(\boldsymbol{y}_1, \boldsymbol{y}_2 |\mathcal{H}_0)$.
Assuming $p(\mathcal{H}_1) = p(\mathcal{H}_0) = \frac{1}{2}$, the  probability for a same-author trial is calculated as~\cite{boenninghoff:2020a}:
\begin{align}
 \label{eq:llrs}
  p_{\text{BFS}}(\mathcal{H}_1|\boldsymbol{y}_1, \boldsymbol{y}_2)
        = \frac{p(\boldsymbol{y}_1, \boldsymbol{y}_2|\mathcal{H}_1)}
            {p(\boldsymbol{y}_1, \boldsymbol{y}_2|\mathcal{H}_1) + p(\boldsymbol{y}_1, \boldsymbol{y}_2|\mathcal{H}_0)}
        = \text{Sigmoid}\big( \text{score}(\boldsymbol{y}_1, \boldsymbol{y}_2) \big)
\end{align}

\subsubsection{Loss function:}
The calculation of Eq.~\eqref{eq:llrs} requires numerically stable inversions of matrices~\cite{boenninghoff:2020a}.
Hence, we firstly reduce the dimension of the \gls{LEVs} via
\begin{align}
\label{eq:nllbfs}
    \boldsymbol{y}_i^{\textsc{BFS}} = f^{\textsc{BFS}}\big(\boldsymbol{W}^{\textsc{BFS}} \boldsymbol{y}_i +  \boldsymbol{b}^{\textsc{BFS}}\big),
\end{align}
where $f^{\textsc{BFS}}(\cdot)$ represents the chosen activation function  (see Section~\ref{sec:EXP}).
We rewrite Eq.~\eqref{eq:llrs} as follows
\begin{align}
 \label{eq:scorefinal} 
  p_{\text{BFS}}(\mathcal{H}_1|\boldsymbol{y}_1, \boldsymbol{y}_2)
        = \text{Sigmoid}\big( \text{score}(\boldsymbol{y}_1^{\textsc{BFS}}, \boldsymbol{y}_2^{\textsc{BFS}}) \big)
\end{align}
and incorporate Eq.~\eqref{eq:scorefinal} into the binary cross entropy,
\begin{align}
\label{eq:loss2}
\loss^{\text{BFS}}_{\boldsymbol{\phi}} 
    = a \cdot \log \left\{ p_{\text{BFS}}(\mathcal{H}_1|\boldsymbol{y}_1, \boldsymbol{y}_2) \right\}
            + (1-a) \cdot \log \left\{ 1 - p_{\text{BFS}}(\mathcal{H}_1|\boldsymbol{y}_1, \boldsymbol{y}_2) \right\},
\end{align}
where all trainable parameters are denoted with $\boldsymbol{\phi} = \big\{\boldsymbol{W}^{\textsc{BFS}}, \boldsymbol{b}^{\textsc{BFS}}, \boldsymbol{W}, \boldsymbol{B}, \boldsymbol{\mu} \big\}$.
We also consider the within-author and between-authors variabilities by determining the Gaussian entropy
during training. As shown in Fig.~\ref{fig:loss}, the within-author variability decreases while the between-author variability increases.

\subsection{Uncertainty Modeling and Adaptation}
We expect that the \gls{BFS} component returns a mixture of correct and mislabelled trials. 
We therefore treat the posteriors as noisy outcomes and rewrite Eq.~\eqref{eq:scorefinal} as
$p_{\text{BFS}}(\widehat{\mathcal{H}}_1|\boldsymbol{y}_1, \boldsymbol{y}_2)$
to emphasize that this represents an estimated posterior.
Inspired by~\cite{luo-etal-2017-learning-noise}, the idea is to find wrongly classified trials and to model the noise behavior of the \gls{BFS}. 
We firstly have to find a single representation for both \gls{LEVs}, which is done by
\begin{align}
    \boldsymbol{y}^{\textsc{UAL}} = \tanh\big(\boldsymbol{W}^{\textsc{UAL}} 
    \big( \boldsymbol{y}_1 - \boldsymbol{y}_2 \big)^{\circ 2} +  \boldsymbol{b}^{\textsc{UAL}}\big),
\end{align}
where 
$(\cdot)^{\circ 2}$ denotes the element-wise square. 
Next, we compute a $2\times 2$ confusion matrix as follows
\begin{align}
    \label{eq:ualC}
    p(\mathcal{H}_j | \widehat{\mathcal{H}}_i, \boldsymbol{y}_1, \boldsymbol{y}_2) =  \frac{\exp\big( \boldsymbol{w}_{ji}^T ~\boldsymbol{y}^{\textsc{BFS}}  + b_{ji} \big)}
{\sum\limits_{i'\in\{0,1\}}  \exp\big(\boldsymbol{w}_{ji'}^T ~ \boldsymbol{y}^{\textsc{BFS}} + b_{ji'} \big)}
\quad \text{ for } i,j\in \{0,1\}.
\end{align}
The term $ p(\mathcal{H}_j | \widehat{\mathcal{H}}_i, \boldsymbol{y}_1, \boldsymbol{y}_2)$ defines the conditional probability of the true hypothesis $\mathcal{H}_j$ given 
the assigned hypothesis $\widehat{\mathcal{H}}_i$ by the \gls{BFS}.
Here, vector $\boldsymbol{w}_{ji}$ and bias term $b_{ji}$ characterize the confusion between $j$ and $i$. 
We can then adapt the uncertainty to define the final output predictions:
\begin{align}
\label{eq:UAL}
p_{\text{UAL}}(\mathcal{H}_j| \boldsymbol{y}_1, \boldsymbol{y}_2)
 = \sum\limits_{i\in \{0,1\}}    
        p(\mathcal{H}_j | \widehat{\mathcal{H}}_i, \boldsymbol{y}_1, \boldsymbol{y}_2)
        \cdot p_{\text{BFS}}(\widehat{\mathcal{H}}_i | \boldsymbol{y}_1, \boldsymbol{y}_2).
\end{align}

\subsubsection{Loss function:}
The loss consists of two terms, the negative log-likelihood of the groundtruth hypothesis and a regularization term,
\begin{align}
\label{eq:lossual}
 \begin{split}
\loss^{\text{UAL}}_{\boldsymbol{\lambda}} 
   & = -\log p_{\text{UAL}}(\mathcal{H}_j| \boldsymbol{y}_1, \boldsymbol{y}_2) 
   \\&\quad + \beta~ \sum_{i\in\{0,1\}} \sum_{j\in \{0,1\}}
    p(\mathcal{H}_j | \widehat{\mathcal{H}}_i, \boldsymbol{y}_1, \boldsymbol{y}_2) 
    \cdot
    \log p(\mathcal{H}_j | \widehat{\mathcal{H}}_i, \boldsymbol{y}_1, \boldsymbol{y}_2),
\end{split}
\end{align}
with trainable parameters denoted by $\boldsymbol{\lambda} = \big\{\boldsymbol{W}^{\textsc{UAL}}, \boldsymbol{b}^{\textsc{UAL}}, \boldsymbol{w}_{ji}, \boldsymbol{b}_{ji} | j,i\in \{0,1\}\big\}$.
The regularization term, controlled by $\beta$, follows the maximum entropy principle to penalize the confusion matrix for returning over-confident posteriors~\cite{pereyra2017regularizing}. 
We observed that the probabilities are usually placed closer to zero or one, which is equivalent to a distribution with low entropy.
Without regularization, either $p(\mathcal{H}_0 | \widehat{\mathcal{H}}_0, \boldsymbol{y}_1, \boldsymbol{y}_2) \approx p(\mathcal{H}_0 | \widehat{\mathcal{H}}_1, \boldsymbol{y}_1, \boldsymbol{y}_2) \approx 1 $ or 
$p(\mathcal{H}_1 | \widehat{\mathcal{H}}_1, \boldsymbol{y}_1, \boldsymbol{y}_2) \approx p(\mathcal{H}_1 | \widehat{\mathcal{H}}_0, \boldsymbol{y}_1, \boldsymbol{y}_2) \approx 1 $.
The objective of the maximum entropy regularizer is to reduce this effect.

\subsection{Overall Loss Function:}
The overall loss combines the model accuracy, as assessed in Bayes factor scoring, with the uncertainty adaptation loss:
\begin{align}
\label{eq:lossall}
 \begin{split}
\loss_{\boldsymbol{\theta}, \boldsymbol{\psi},\boldsymbol{\phi},  \boldsymbol{\lambda}} 
 = \loss^{\text{DML}}_{\boldsymbol{\theta}, \boldsymbol{\psi}} 
+ \loss^{\text{BFS}}_{\boldsymbol{\phi}} 
 + \loss^{\text{UAL}}_{\boldsymbol{\lambda}}. 
\end{split}
\end{align}
All components are optimized independently w.r.t.~the corresponding loss.

\section{Experiments}
\label{sec:EXP}
The overall score of the PAN 2021 shared task is given by averaging five metrics~\cite{kestemont:2020}: \texttt{AUC} measures true/false positive rates for various thresholds. \texttt{F1} is defined as the harmonic mean of precision and recall. In this work, the \texttt{c@1} score represents the accuracy, since we do not return non-responses. The \texttt{f\_05\_u} favors systems deciding same-author trials correctly.
Finally, the \texttt{Brier} score rewards systems that return correct and  self-confident predictions.
To capture the calibration capacity, we provide the \texttt{ECE} and \texttt{MCE} metrics.
The first one computes the weighted macro-averaged absolute error between confidence and accuracy of all bins. The latter returns the maximum absolute error~\cite{pmlr-v70-guo17a}.

\begin{figure}[t]
\vspace*{-0.4cm}
\centering
\begin{subfigure}[t]{0.48\textwidth}
\centering
\includegraphics[width=0.99\textwidth]{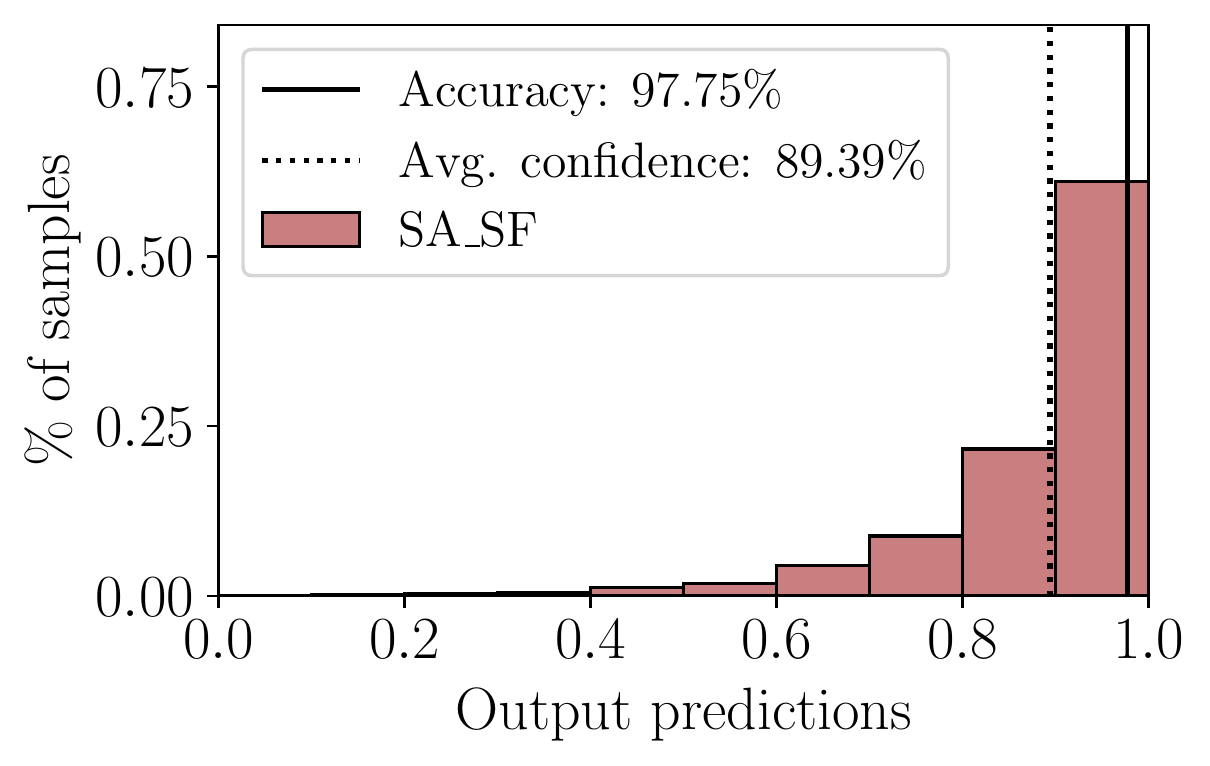}
\caption{Histogram for \texttt{SA\_SF} pairs.}
\end{subfigure}
\quad
\begin{subfigure}[t]{0.48\textwidth}
\centering
\includegraphics[width=0.99\textwidth]{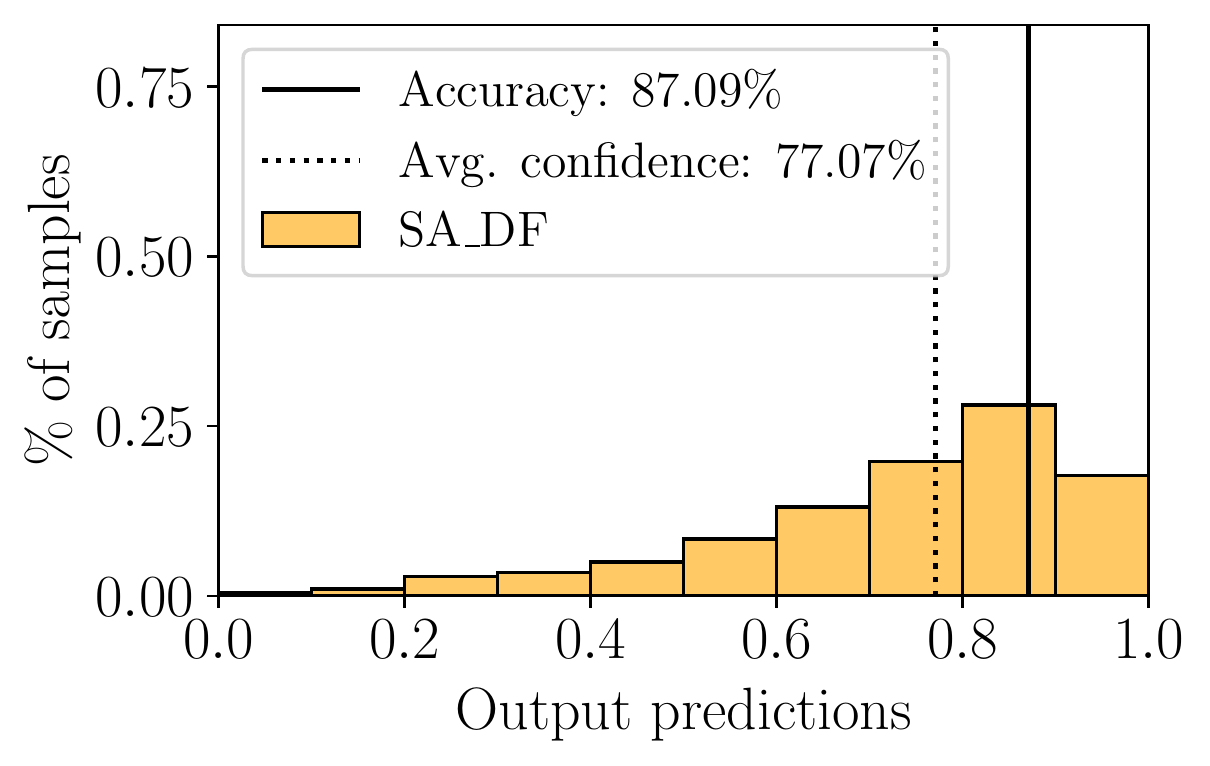}
\caption{Histogram for \texttt{SA\_DF} pairs.}
\end{subfigure}
\begin{subfigure}[t]{0.48\textwidth}
\centering
\includegraphics[width=0.99\textwidth]{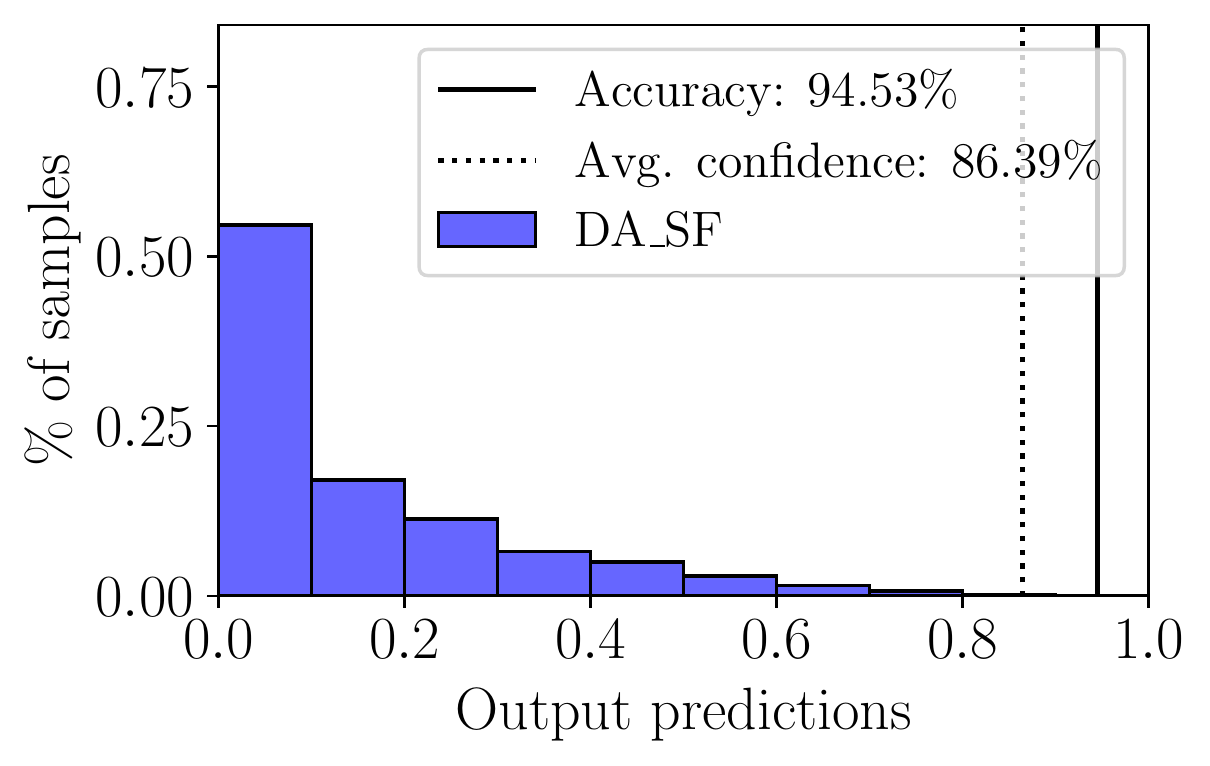}
\caption{Histogram for \texttt{DA\_SF} pairs.}
\end{subfigure}
\quad
\begin{subfigure}[t]{0.48\textwidth}
\centering
\includegraphics[width=0.99\textwidth]{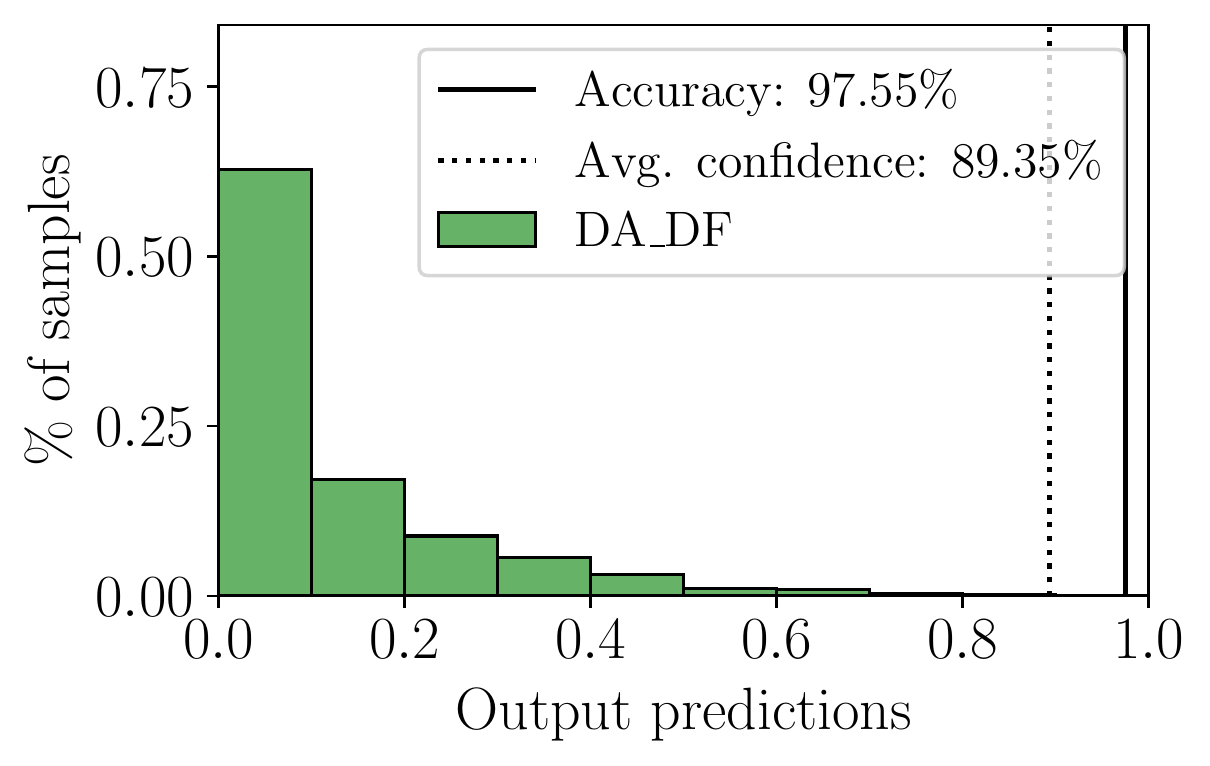}
\caption{Histogram for \texttt{DA\_DF} pairs.}
\end{subfigure}
\vspace*{-0.2cm}
\caption{Posterior histograms for \gls{DML} with fixed kernel parameters.} 
\label{fig:cal_his_all}
\end{figure}

\begin{figure}[t]
\vspace*{-0.4cm}
\centering
\begin{subfigure}[t]{0.48\textwidth}
\centering
\includegraphics[width=0.99\textwidth]{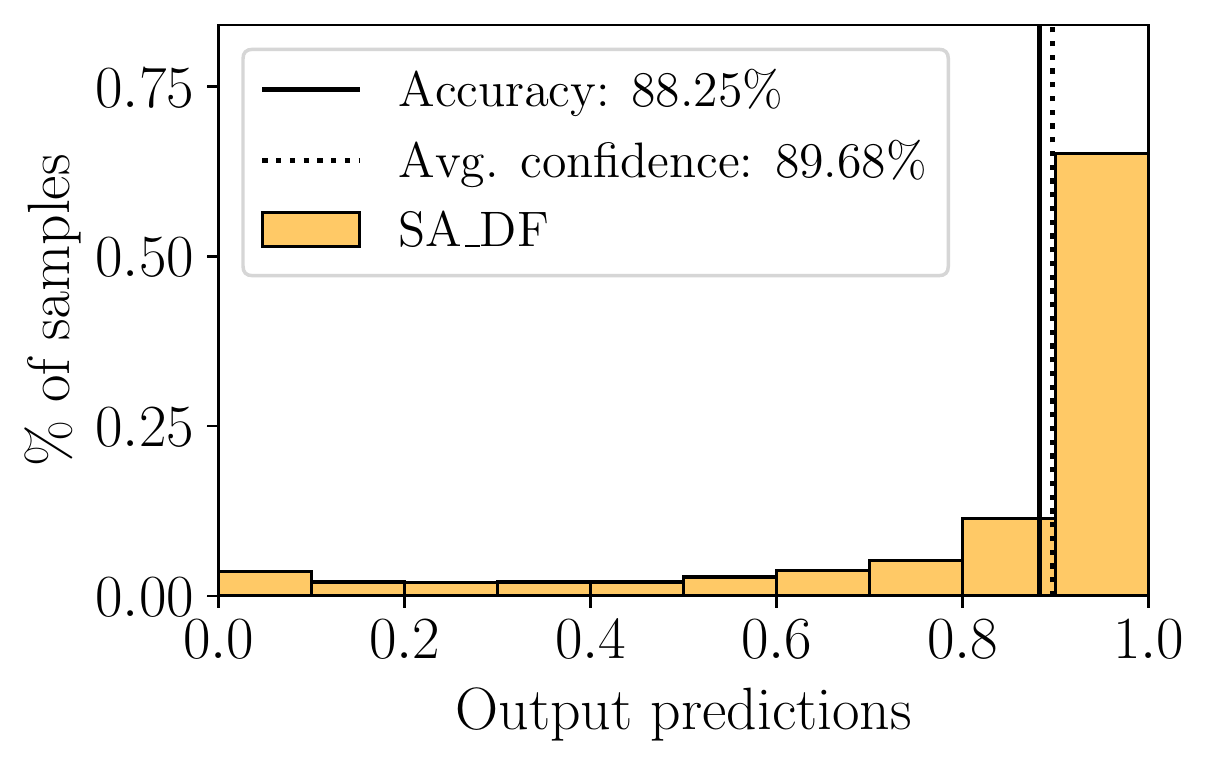}
\caption{Histogram for \texttt{SA\_DF} pairs.}
\end{subfigure}
\quad
\begin{subfigure}[t]{0.48\textwidth}
\centering
\includegraphics[width=0.99\textwidth]{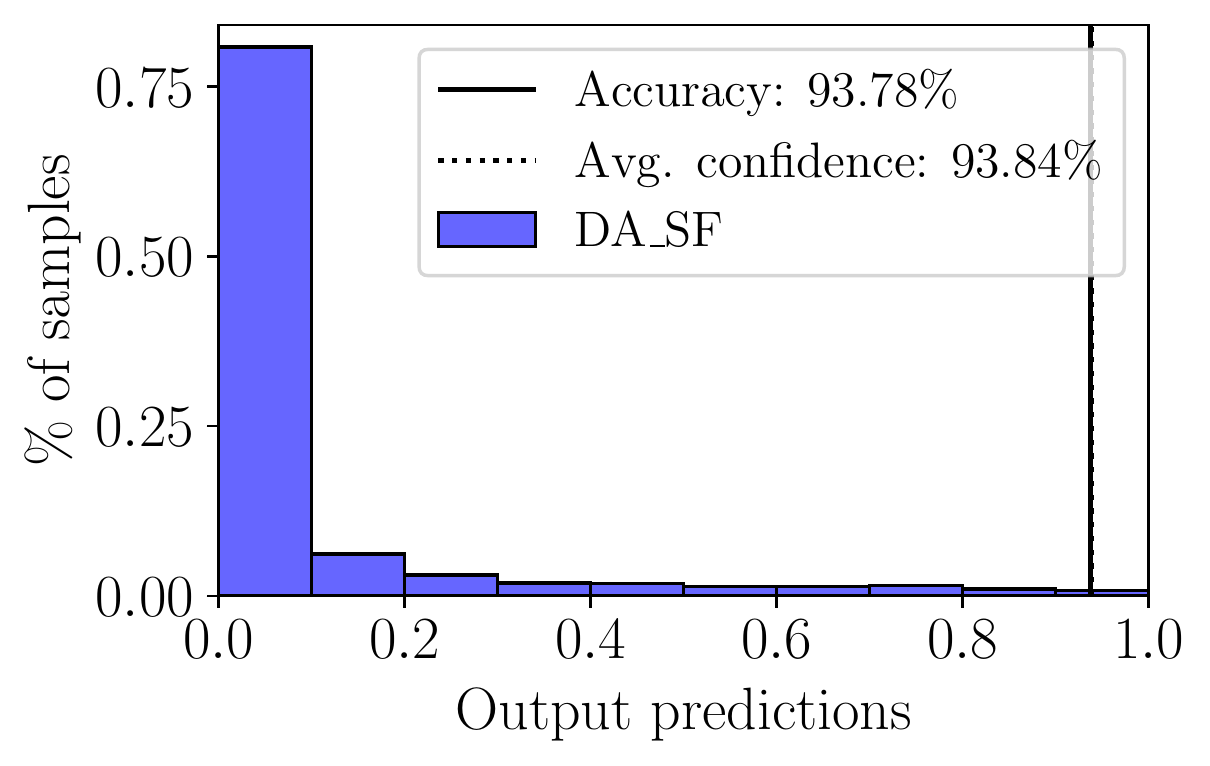}
\caption{Histogram for \texttt{DA\_SF} pair.}
\end{subfigure}
\vspace*{-0.2cm}
\caption{Posterior histograms after uncertainty adaptation ($\beta=0.1$).} 
\label{fig:cal_his_ual}
\end{figure}

\subsection{Analysis of the Sensitivity to Topical Interference}
In a first step, we evaluated the discriminative power of the \gls{DML} component alone, for fixed parameters $\alpha$ and $\gamma$ as described in Eq.~\eqref{eq:DMLl}. In Fig.~\ref{fig:cal_his_all}, we show histograms of the posteriors including the accuracy and averaged confidence for a single run.
All confidence values (also in Fig.~\ref{fig:rel2021}) lie within the interval $[0.5, 1]$, since we solve a binary classification task. 
Hence, to obtain confidence scores, the posterior values are transformed w.r.t.~to the estimated authorship label, showing
$p(\mathcal{H}_1|\boldsymbol{y}_1, \boldsymbol{y}_2)$ if $\widehat{a}=1$ and $1 - p(\mathcal{H}_1|\boldsymbol{y}_1, \boldsymbol{y}_2)$ if $\widehat{a}=0$.

It can be clearly seen that all subsets exhibit a high degree of miscalibration, where the accuracy is significantly larger than the corresponding confidence. For all subsets, the \text{DML} model tends to be under-confident in its predictions.

It is also worth noting that topical interference leads to lower performance. Comparing the histograms in Fig.~\ref{fig:cal_his_all} (c) and (d), the accuracy of \texttt{DA\_SF} pairs is 3\% lower than the accuracy of \texttt{DA\_DF} pairs. For \texttt{SA\_SF} and \texttt{SA\_DF} pairs in Fig.~\ref{fig:cal_his_all} (a) and (b), the topical interference is even more obvious and the histogram of \texttt{SA\_DF} pairs almost resembles a uniform distribution.
Fig.~\ref{fig:cal_his_ual} displays the confidence histograms after the proposed uncertainty adaptation layer. We can observe a \textit{self-calibrating} effect, where confidence is much
better aligned with accuracy on average. However, plot (a) in Fig.~\ref{fig:cal_his_ual} also reveals that the model returns a small number of self-confident but wrongly classified same-author trials. 

Our most important discovery at this point is that our model analyzes different-authors pairs more readily. 
As illustrated in Fig~\ref{fig:zipf}, this can be explained by the difficulty of re-sampling heterogeneous subsets of same-authors trials. 
Experiments conducted on a large dataset of Amazon reviews in~\cite{boenninghoff:2019b}, where we had to limit the total contribution of each author, have shown lower error rates for same-authors pairs.

\subsection{Ablation Study}
Next, we focus on the more problematic \texttt{SA\_DF} and \texttt{DA\_SF} cases. All model components are analyzed separately to illustrate notable effects of some hyper-parameters.
We observed that all runs generally achieved best results between epochs 29 and 33. To avoid cherry-picking, we averaged the metrics over these epochs and at least over four runs totally.
All PAN metrics and the corresponding calibration scores are summarized in Table~\ref{tab:results:pan}. Here, we also provide the averaged confidence, allowing us to characterize a system as over- (\texttt{c@1}$<$\texttt{conf}) or under-confident (\texttt{c@1}$>$\texttt{conf}). 

In the first two rows, we see the performance of the \gls{DML} component, showing the effect of learning the kernel parameters $\alpha$ and $\gamma$. The overall score slightly increases, which mainly follows from a better \texttt{Brier} score. As can be seen in Fig.~\ref{fig:mapping}, the learned mapping holds the distance $d(\boldsymbol{y}_1, \boldsymbol{y}_2)$ of a pair close to one or zero over a wider range, resulting in significantly reduced calibration errors. 

The next two rows 
provide the results of the \gls{BFS} component for two different activation functions $f^{\rm BFS}$ in Eq.~\eqref{eq:nllbfs}.
We tried some variations of the $\texttt{ReLU}$ function but did not notice any performance differences and finally proceeded with the $\texttt{Swish}$ function~\cite{DBLP:journals/corr/abs-1710-05941}.
The \texttt{ECE} and \texttt{MCE} show further significant improvements for both activation functions and the overall score slightly increases using the $\texttt{Swish}$ activation. 
Comparing the \texttt{c@1}, \texttt{f\_05\_u} and \texttt{F1} scores, it is noticeable that the choice of activation function can clearly influence the performance metrics.

The last six rows provide a comparison of the \gls{UAL} component\footnote{$\alpha,\gamma$ are learned, first four rows for \texttt{tanh} activation, last two rows for \texttt{Swish} activation.}. 
The fourth row shows that the output of \gls{BFS} returns slightly over-confident predictions (\texttt{c@1}$<$\texttt{conf}). The \gls{UAL} without regularization $(\beta=0)$ only reinforces this trend. 
We varied the parameter $\beta$ over the range $[0.05, 0.1, 0.125, 0.2]$.
We observed that increasing $\beta$ generally reduces the over-confidence of the model and with $\beta=0.2$ in the 8th row, the influence of the regularizer becomes so strong that output predictions are now under-confident. In addition, the higher the $\beta$ parameter is chosen, the lower the \texttt{MCE} values become for \texttt{tanh} activation while it remains on the same level for the \texttt{Swish} activation.
This offers a mechanism to optimize the calibration, decreasing the \texttt{ECE} to approximately $0.7-0.8\%$.

\subsection{Discussion}

Our experiments yield two findings: 
First, as intended, the Bayes factor scoring together with uncertainty adaptation and maximum entropy regularization achieve a high agreement
between model confidence and accuracy.
Secondly, we are able to slightly increase the overall performance score.
Our results from the PAN 2020 shared task, furthermore, show that our framework can better capture the writing style of a person compared to traditional hand-crafted features or compression-based approaches. 

\begin{table}[t]
\vspace*{-0.23cm}
\centering
\renewcommand{\arraystretch}{1.3}
\caption{Results for PAN 2021 evaluation and calibration metrics.}
\label{tab:results:pan}
\resizebox{\textwidth}{!}{
    \begin{tabular}{|c c | c |c |c |c |c |c|c|c|c|}
    \hline
   \multicolumn{2}{|c|}{\multirow{2}{*}{\textbf{Model}}}
    & \multicolumn{6}{c|}{\textbf{PAN 2021 Evaluation Metrics}}
    & \multicolumn{3}{c|}{\textbf{Calibration Metrics}}
       \\ \cline{3-11}
    & &\texttt{~AUC~}           &\texttt{~c@1~}       &\texttt{f\_05\_u}    &\texttt{~F1~}  &\texttt{~Brier~}      &\texttt{overall} 
    &\texttt{~conf~}   
    &\texttt{~ECE~}           
    &\texttt{~MCE~}
    \\ \hline
    \multirow{2}{*}{\begin{adjustbox}{angle=90}\textbf{DML}\end{adjustbox}} 
    & \multicolumn{1}{|c|}{fixed}
    &$97.3 \pm 0.1$  
    &$91.2 \pm 0.2$  
    &$\boldsymbol{91.7} \pm \boldsymbol{0.2}$  
    &$89.2 \pm 0.2$ 
    &$92.4 \pm 0.1$ 
    &$92.4 \pm 0.1$
    &$82.9 \pm 0.1$
    &$7.9 \pm 1.0$ 
    &$16.6 \pm 0.5$
        \\ 
    & \multicolumn{1}{|c|}{learned} 
    &$97.3 \pm 0.1$  
    &$91.5 \pm 0.2$  
    &$90.6 \pm 0.7$  
    &$89.9 \pm 0.3$ 
    &$93.5 \pm 0.3$ 
    &$92.6 \pm 0.2$ 
    &$89.4 \pm 0.2$
    &$2.2 \pm 0.2$ 
    &$~~7.9 \pm 1.0$
    \\ \hline
     \multirow{2}{*}{\begin{adjustbox}{angle=90}\textbf{BFS}
     \end{adjustbox}}
     &\multicolumn{1}{|c|}{\texttt{Swish}}
    &$\boldsymbol{97.4} \pm \boldsymbol{0.1}$  
    &$\boldsymbol{91.6} \pm \boldsymbol{0.2}$  
    &$90.9 \pm 0.4$  
    &$89.9 \pm 0.3$ 
    &$93.7 \pm 0.1$ 
    &$92.7 \pm 0.1$ 
    &$91.4 \pm 0.2$
    &$0.8 \pm 0.1$ 
    &$~~4.4\pm 1.3$
        \\ 
    & \multicolumn{1}{|c|}{\texttt{tanh}} 
    &$97.3 \pm 0.1$  
    &$91.2 \pm 0.2$  
    &$91.5 \pm 0.3$  
    &$89.2 \pm 0.4$ 
    &$93.4 \pm 0.2$ 
    &$92.5 \pm 0.1$ 
    &$91.8 \pm 0.1$
    &$1.1 \pm 0.2$ 
    &$~~4.7 \pm 1.5$
     \\ \hline
     \multirow{4}{*}{\begin{adjustbox}{angle=90}\textbf{UAL} (\texttt{tanh})
     \end{adjustbox}}
 & \multicolumn{1}{|c|}{$\beta=0$} 
    &$97.3 \pm 0.1$  
    &$91.5 \pm 0.3$  
    &$91.4 \pm 0.4$  
    &$89.8 \pm 0.5$ 
    &$93.6 \pm 0.3$ 
    &$92.7 \pm 0.3$ 
    &$93.7 \pm 0.2$
    &$2.3 \pm 0.4$ 
    &$~~8.1 \pm 1.5$
    \\
& \multicolumn{1}{|c|}{$\beta=0.05$} 
    &$97.3 \pm 0.1$  
    &$91.4 \pm 0.2$  
    &$91.0 \pm 0.4$  
    &$89.7 \pm 0.3$ 
    &$93.7 \pm 0.1$ 
    &$92.6 \pm 0.1$ 
    &$92.7 \pm 0.2$
    &$1.4 \pm 0.2$ 
    &$~~6.7 \pm 1.7$
     \\
& \multicolumn{1}{|c|}{$\beta=0.1$} 
    &$97.3 \pm 0.1$  
    &$91.5 \pm 0.2$  
    &$91.2 \pm 0.4$  
    &$89.8 \pm 0.2$ 
    &$93.7 \pm 0.1$ 
    &$92.7 \pm 0.2$
    &$92.1 \pm 0.2$
    &$0.8 \pm 0.1$ 
    &$~~5.4 \pm 1.5$    \\
& \multicolumn{1}{|c|}{$\beta=0.2$} 
    &$\boldsymbol{97.4} \pm \boldsymbol{0.1}$  
    &$91.5 \pm 0.2$  
    &$91.1 \pm 0.4$  
    &$89.8 \pm 0.3$ 
    &$93.7 \pm 0.1$ 
    &$92.7 \pm 0.1$ 
    &$90.2 \pm 0.2$
    &$1.6 \pm 0.2$ 
    &$~~\boldsymbol{4.1} \pm \boldsymbol{1.1}$
    \\ \hline
     \multirow{2}{*}{\begin{adjustbox}{angle=90}\textbf{UAL}
     \end{adjustbox}}
 & \multicolumn{1}{|c|}{$\beta=0.1$} 
    &$97.3 \pm 0.0$  
    &$91.5 \pm 0.1$  
    &$90.9 \pm 0.4$  
    &$89.9 \pm 0.2$  
    &$\boldsymbol{93.8} \pm \boldsymbol{0.1}$  
    &$92.7 \pm 0.1$  
    &$91.9 \pm 0.2$  
    &$\boldsymbol{0.7} \pm  \boldsymbol{0.1}$  
    &~~$4.8 \pm 1.3$  
    \\
& \multicolumn{1}{|c|}{$\beta=0.125$} 
    &$\boldsymbol{97.4} \pm \boldsymbol{0.1}$  
    &$\boldsymbol{91.6} \pm \boldsymbol{0.1}$  
    &$91.0 \pm 0.3$  
    &$\boldsymbol{90.0} \pm \boldsymbol{0.1}$  
    &$\boldsymbol{93.8} \pm \boldsymbol{0.1}$  
    &$\boldsymbol{92.8} \pm \boldsymbol{0.1}$  
    &$91.6 \pm 0.1$  
    &$0.8 \pm  0.1$  
    &~~$5.0 \pm 1.6$  
    \\ \hline
    \end{tabular}
}
\end{table}

Nevertheless, our model is constrained by limits in the discriminative power of the employed \gls{LEVs},
which serve as the input to all of the subsequent components. 
One critical point is that LEVs may only capture the surface structure of the writing style. The visualization of the attentions in~\cite{boenninghoff:2019b} shows that the system primarily focuses on easily identifiable features, like orthography or punctuation. 
Another issue is the use of the chosen word representations, which are limited to represent the semantic meaning of a word only in a small context.

We can further improve our framework by addressing the two major types of uncertainty~\cite{NIPS2017_2650d608}: 
On the one hand, \textit{aleatoric} or data uncertainty is associated with properties of the document pairs and captures noise inherent in each document.
Examples are topical variations, the intra- and inter-author variabilities or the varying lengths of documents. Aleatoric uncertainty generally can not be reduced, even if more training pairs become available, but it can be learned along with the model. 
Aleatoric uncertainty can be captured by returning a non-response, when it is hard to decide for one hypothesis $\mathcal{H}_0$ or $\mathcal{H}_1$. 

On the other hand, \textit{epistemic} or model uncertainty characterizes uncertainty in the model parameters.
Examples are the lack of knowledge, e.g. out-of-distribution document pairs or the described issue of re-sampling heterogeneous same-author pairs. 
This uncertainty obviously can be explained away given enough training pairs.
One way to capture epistemic uncertainty is to extend our model to an ensemble. We expect all models to behave similarly for known authors or topics. But the predictions may be widely dispersed for pairs under covariate shift~\cite{10.5555/3295222.3295387}.
We will discuss our approaches for capturing these uncertainties and defining non-responses in the PAN 2021 submission paper.


\section{Conclusion}
In this work, we present a hybrid neural-probabilistic framework to address the task of authorship verification.
We generally achieve high overall scores under covariate shift and we further show that our framework mitigates two  fundamental  problems: topic variability and miscalibration.

In forensic applications, the requirement exists to return suitable and well-calibrated likelihood-ratios rather than decisions. 
However, in the context of the PAN shared tasks, the evaluation protocol assesses decisions. 
Nevertheless, the experiments show that we are closing in on a well-calibrated system, which would allow us to interpret the obtained confidence score or posterior as the probability of a correct decision and to bridge the gap between computational authorship verification and traditional forensic text comparison.

\section*{Acknowledgment}
\small
This work was in significant parts performed on a HPC cluster at Bucknell University through the support of the National Science Foundation, Grant Number 1659397. Project funding was provided by the state of North Rhine-Westphalia within the Research Training Group "SecHuman - Security for Humans in Cyberspace" and by the Deutsche Forschungsgemeinschaft (DFG) under Germany’s Excellence Strategy - EXC2092CaSa- 390781972.

\small
\bibliographystyle{template/splncs04}
\bibliography{refs}

\end{document}